\documentclass[journal,twoside,web]{ieeecolor}
\usepackage{generic}
\usepackage{color}
\usepackage{graphicx}
\usepackage{cite}
\usepackage{picinpar}
\usepackage{amsmath}
\usepackage{url}
\usepackage{flushend}
\usepackage[latin1]{inputenc}
\usepackage{colortbl}
\usepackage{soul}
\usepackage{multirow}
\usepackage{pifont}
\usepackage{alltt}
\usepackage[hidelinks]{hyperref}
\usepackage{enumerate}
\usepackage{siunitx}
\usepackage{breakurl}
\usepackage{epstopdf}
\usepackage{pbox}
\usepackage{amssymb}
\usepackage{subfigure}
\usepackage{pifont}
\usepackage{booktabs}
\usepackage[switch]{lineno}
\usepackage{makecell}

\usepackage{rotating}

\begin{document}

\title{An Efficient MLP-based Point-guided Segmentation Network for Ore Images with Ambiguous Boundary}
\author{Guodong~Sun, Yuting~Peng, Le~Cheng, Mengya Xu, An Wang, Bo Wu, Hongliang Ren, and Yang~Zhang
\thanks{Corresponding author: Yang~Zhang.}
\thanks{G. Sun, Y. Peng, L. Cheng, and Y. Zhang are with the School of Mechanical Engineering, Hubei University of Technology, Wuhan 430068, China (e-mail: sunguodong@hbut.edu.cn; pyt181@hbut.edu.cn; cl@hbut.edu.cn; yzhangcst@hbut.edu.cn ).}
\thanks{ M. Xu and H. Ren are with the Department of Biomedical Engineering, National University of Singapore (NUS), Singapore 117575, Singapore (e-mail: mengya@u.nus.edu; hlren@ieee.org).}
\thanks{ A. Wang, H. Ren, and Y. Zhang are also with the Department of Electronic Engineering, The Chinese University of Hong Kong, Hong Kong 999077, China (e-mail: wa09@link.cuhk.edu.hk).}
\thanks{B. Wu is with the Shanghai Advanced Research Institute, Chinese Academy of Sciences, Shanghai 201210, China (e-mail: wubo@sari.ac.cn).}
}

\maketitle

\begin{abstract}
The precise segmentation of ore images is critical to the successful execution of the beneficiation process. Due to the homogeneous appearance of the ores, which leads to low contrast and unclear boundaries, accurate segmentation becomes challenging, and recognition becomes problematic. This paper proposes a lightweight framework based on Multi-Layer Perceptron (MLP), which focuses on solving the problem of edge burring.  Specifically, we introduce a lightweight backbone better suited for efficiently extracting low-level features. Besides, we design a feature pyramid network consisting of two MLP structures that balance local and global information thus enhancing detection accuracy. Furthermore, we propose a novel loss function that guides the prediction points to match the instance edge points to achieve clear object boundaries. We have conducted extensive experiments to validate the efficacy of our proposed method. Our approach achieves a remarkable processing speed of over 27 frames per second (FPS) with a model size of only 73 MB. Moreover, our method delivers a consistently high level of accuracy, with impressive performance scores of 60.4 and 48.9 in~$AP_{50}^{box}$ and~$AP_{50}^{mask}$ respectively, as compared to the currently available state-of-the-art techniques, when tested on the ore image dataset. The source code will be released at \url{https://github.com/MVME-HBUT/ORENEXT}.
\end{abstract}


\definecolor{limegreen}{rgb}{0.2, 0.8, 0.2}
\definecolor{forestgreen}{rgb}{0.13, 0.55, 0.13}
\definecolor{greenhtml}{rgb}{0.0, 0.5, 0.0}

\section{Introduction}

\begin{figure}[!t]
	\centering
	
	\subfigure[Input]{
		\begin{minipage}[c]{0.2\linewidth}
			\includegraphics[width=0.8in]{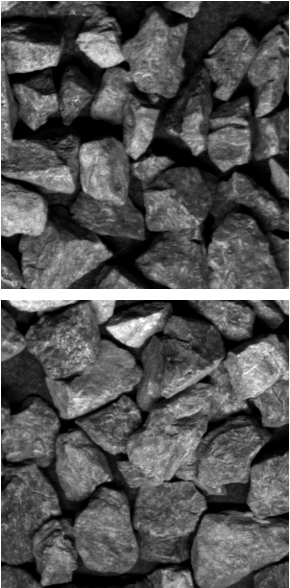}\vspace{0.4em} 
		\end{minipage}
	 }\hspace{0.1em}
    \subfigure[GroundTruth]{
		\begin{minipage}[c]{0.2\linewidth}
			\includegraphics[width=0.8in]{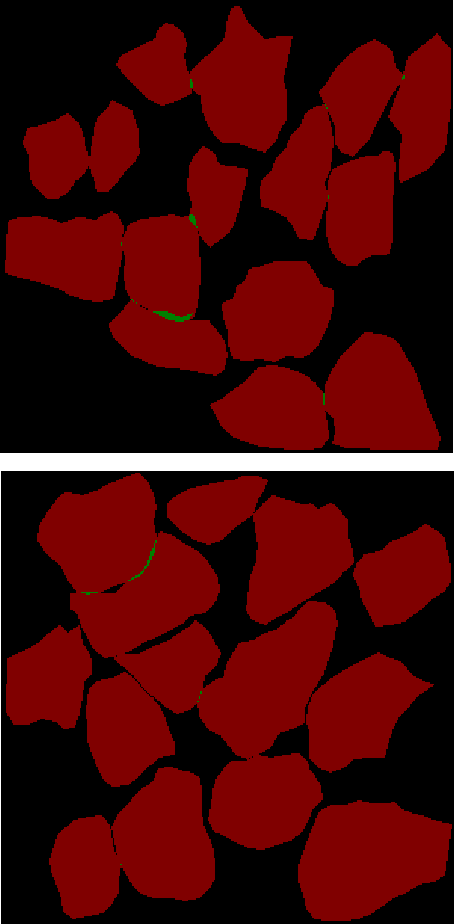}\vspace{0.4em} 
		\end{minipage}
    }\hspace{0.1em}
	\subfigure[Baseline]{
		\begin{minipage}[c]{0.2\linewidth}
			\includegraphics[width=0.8in]{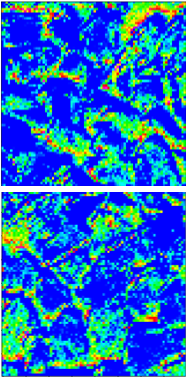}\vspace{0.4em} 
		\end{minipage}
	}\hspace{0.1em}
	\subfigure[Ours]{
		\begin{minipage}[c]{0.2\linewidth}
			\includegraphics[width=0.8in]{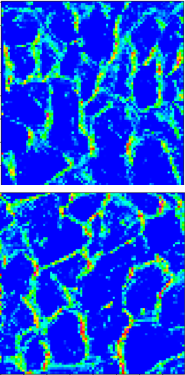}\vspace{0.4em} 	
		\end{minipage}
	}
	\caption{Visualizations of the feature maps. The ore stacking in the input image makes the boundary difficult to distinguish. The edge features of the feature maps obtained by the baseline are blurred. The consistency instances with clearer edges can be obtained using our network.}

  \label{visual}
\end{figure}

\IEEEPARstart{O}{re} particle size analysis is an important part of ore processing tasks. Particle size information is an essential indicator to judge the effectiveness of the crusher, and it can serve as a guide for adjusting the process parameters of each process. Accurate ore segmentation is a significant prerequisite for particle size statistical analysis. 
The production site for beneficiation presents a complex environment with several challenging factors, such as ore stacking, ore adhesion resulting from dry-wet mixing, and variations in lighting conditions, as depicted in \textcolor{red}{Fig.~\ref{visual}}. These factors pose a significant challenge to achieving accurate ore segmentation.
Moreover, the beneficiation scene is often arranged in the field, so the limitation of equipment resources is also a problem faced by the ore particle size analysis system.

Traditional image processing methods perform the segmentation of ore by setting thresholds, cluster analysis, and edge detection. With the development of convolutional neural networks (CNN), the algorithms based on deep learning have shown significant advancements in automatic feature extraction and generalization performance{~\cite{tii1,tii3,tii5,review2,pr}}, so they have gradually gained a leading position in industrial image processing. Most existing instance segmentation methods are implemented on public datasets such as MS COCO~\cite{coco}. These methods are not effective for ore image processing.   It can be seen from \textcolor{red}{Fig.~\ref{visual}(c)} that the baseline framework is less capable of segmenting the ore edges. The processing of ore images is significantly challenged by complex working environments and the presence of diverse feature information.
Aiming at the specific needs of ore segmentation, the U-Net was used to segment broken stones for the first time. However, these complex CNN-based methods have high computational costs, large model sizes, and low accuracy. Moreover, facing the complex environment, the existing algorithmic framework will easily ignore the problem of edge blur caused by inter-adhesive and shadowing of ore images. In this work, we focus on solving this problem and designing an efficient network with less computational overhead, fewer parameters, faster inference time, and better performance. 

Recently, it has been found that MLP-based architectures can achieve comparable results to CNN and Transformer methods with less computation. The MLP-Mixer~\cite{mlpmixer} proposed token mixing and channel mixing MLP to allow interaction between spatial locations and channels. The ResMLP~\cite{resmlp} used cross-patch and cross-channel sublayers as components. Inspired by these works, we propose OreNeXt, an instance segmentation model based on the MLP framework for the ore edge problem. A lightweight MLP backbone network for feature extraction is introduced, followed by a feature pyramid network using a SparseMLP module to enhance the semantic information, then we introduce a loss function guided by the edge. Using MLP and a simple hybrid mechanism, we obtain a lightweight model suitable for deployment.

To implement this model framework to solve the above problems, we use the two-stage detector PointRend~\cite{pointrend} as the baseline for the ore task.
For ore tasks, the low-level features of ore overlapping edges are extremely crucial. Therefore, we propose a novel backbone network StoneMLP, which incorporates shifting operations to extract local information corresponding to different axial shifts.
A sparse feature pyramid network is proposed to strengthen the small target information that is easy to ignore and misjudge while maintaining sufficient semantic information. Furthermore, we propose a loss, including detection loss that predicts the foreground score of each point and mask loss that performs edge guidance by dynamically adjusting vertex pairing. As illustrated in \textcolor{red}{Fig.~\ref{visual}(d)}, our proposed framework is more likely to obtain clear ore segmentation results. Experiments on the ORE image dataset demonstrate that our framework performs state-of-the-art methods with smaller model size and faster inference speed. Our main contributions are summarized as follows.
\begin{enumerate}[1)]

    \item { A lightweight image segmentation network OreNeXt is designed, which solves the problem of blurred edges in ores by guiding the boundary points.}
    \item {We propose a lightweight backbone StoneMLP for capturing local correlation and introduce a semantic enhancement SparseFPN network.}
    \item {A loss function that matches edges by guided points is introduced, significantly improving the quality of predicted boundary details.}
    \item{The experimental results validate the efficacy of our method in enhancing the performance of ore image segmentation tasks, while simultaneously reducing the number of parameters and improving inference speed.}
\end{enumerate}

The remainder of this paper is structured as follows. Section~\ref{sec:related_works}  introduces an overview of existing mineral image segmentation methods and MLP architecture development. Section~\ref{sec:method}  presents our overall framework, including two new structures and a new loss function. To validate the effectiveness of our method, in Section~\ref{sec:experiments}, we conduct integrated experiments. The paper is concluded in Section~\ref{sec:conclusion}.

\section{Related works}\label{sec:related_works}

\subsection{Ore image segmentation}
Accurate individual ore segmentation is a crucial prerequisite for granularity statistics and an essential component in ore processing tasks. With the development of CNN, the task of ore image processing has gradually shifted from traditional image processing techniques to deep learning-based segmentation algorithms. For instance, Sun et al.~\cite{hdl} proposed an efficient instance segmentation algorithm to split ore. Dipti et al.~\cite{Dipti} proposed an image segmentation system for estimating the granularity of oil sandstones. The ore image on the conveyor belt often suffers from ore adhesion and occlusion caused by dust and soil. These factors result in the blurring or even complete disappearance of multiple ore edges in the image, posing significant challenges to achieving accurate individual ore segmentation. To address this edge problem, numerous models have been employed to tackle the issue of under-segmentation in ore processing~\cite{chengle}, Li et al.{~\cite{li}} proposed a model based on U-Net, which alleviated the problem of ore granularity detection by improving the loss function and using watershed technology. Liu et al.{~\cite{TIE}} developed the RLPNet to obtain high-precision segmentation images by reducing the interference of complex textures and enhancing the expression of edge features. With the effective emergence of Transformer, numerous Transformer-based designs have been proposed to effectively divide the adhesive ore{~\cite{oretrans2}}, but the network structures are often complex and not suitable for outdoor deployment. Although these methods have improved the segmentation of ore images, there are still problems in clear edge segmentation.

\subsection{Instance segmentation}
Currently, instance segmentation methods can be divided into two categories: one-stage and two-stage. One-stage methods mainly involve a simple fully convolutional networks (FCN) architecture for mask prediction without region of interest (RoI) pooling, such as YOLACT~\cite{YOLACT} and BlendMask~\cite{BlendMask}, proposal generation and feature re-pooling, with fast speed but poor accuracy. Two-stage instance segmentation methods first detect boundary boxes and then perform segmentation in each RoI region. Mask RCNN~\cite{maskrcnn} adds a mask branch to the Faster R-CNN~\cite{Faster-RCNN} architecture. Mask-Refined R-CNN{~\cite{review1}} refines the detailed information of the target and considers the relationship between the edge pixels of the object. PointRend~\cite{pointrend} samples feature points with low confidence scores and further improves their labels using a shared MLP. Cascade Mask R-CNN~\cite{Cascade-RCNN} improves segmentation accuracy through cascading methods. Since most two-stage methods use RoI Align operations, there is information loss in the spatial features of large targets, especially at the edges. Therefore, rough target edge prediction is a common problem for two-stage instance segmentation algorithms.
\begin{figure*}[!t]
	\centering
	\includegraphics[width=6.5in]{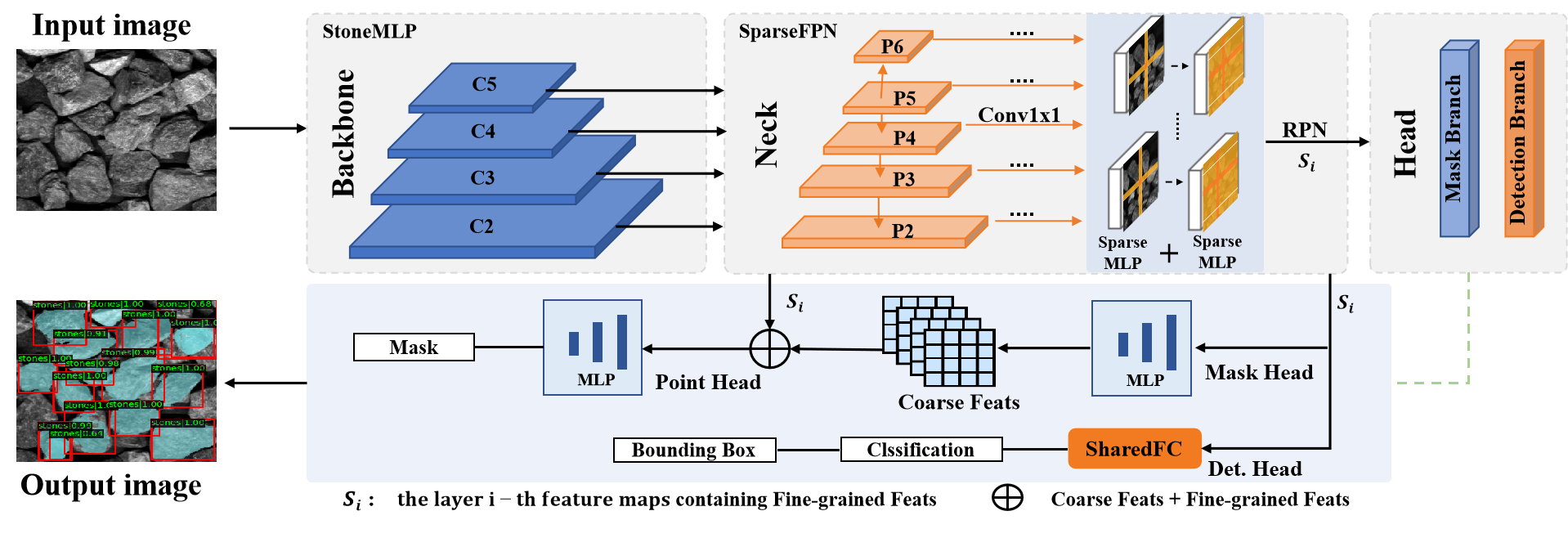}
	\caption{A schematic overview of OreNeXt. The input image is fed into the lightweight backbone StoneMLP (Fig. \ref{StoneMLP}) to produce feature maps. StoneMLP captures local dependencies and extracts edge information through horizontal and vertical shift operations (Fig.~\ref{shift}). Then, the feature maps enter the SparseFPN to generate multi-scale information-integrated feature maps. Our improved FPN structure adds two SparseMLP modules (Fig. \ref{field}), which are divided into three parallel branches for feature fusion through weighted summation. The addition of two sparsely connected SparseMLP modules allows both local and global features to be taken into account. Next, each layer feature map is computationally fused by the region proposal network (RPN) to obtain the fused feature map $S_i$ (the i-th feature maps). Finally,  The point head uses interpolated features computed from the fine-grained feature of the CNN feature maps ($S_i$) and the coarse prediction mask for subdivision prediction.}
	\label{Architecture}
\end{figure*}

\subsection{MLP-Based Models}

With the success of the transformer in the field of natural language processing, some researchers have started exploring how to apply the transformer to the field of vision. ViT~\cite{vit} pioneered the approach of segmenting images into non-overlapping blocks as tokens for operations, enabling the use of the transformer framework for image processing. Swin transformer~\cite{swint} proposed a local group self-attention mechanism incorporating locality. Inspired by the elegant structure of ViT, MLP with simpler network structures has also received a lot of attention in the field of computer vision. MLP-Mixer~\cite{mlpmixer} developed by Google, replaces the self-attention mechanism with MLPs to construct a pure MLP architecture. The ResMLP~\cite{resmlp} used cross-patch and cross-channel sublayers as components. However, most existing MLP-based methods cannot adapt to image resolution, making them difficult to apply to downstream tasks. ASMLP~\cite{asmlp} and CycleMLP~\cite{cyclemlp}, both of which use the motion of feature maps to integrate local information, enabling the pure MLP architecture to be used for downstream tasks. These MLP-based models have a similar overall structure but differ in the detailed design of the main modules. After MLP-Mixer~\cite{mlpmixer} was proposed, people began to explore the architecture of MLP. With increased computational capacity and the availability of larger datasets coupled, the ancient simple structure of MLP can also achieve effective performance improvement in various vision tasks. Through the design structure, these MLP-based models no longer rely on prior knowledge of two-dimensional images, but on the input feature maps for long-range interactions. However, this rarely makes full use of local information, and the extraction of low-level features is also significant for many tasks.

\section{Method}\label{sec:method}
In this section, we initially introduce the overall structure of the network. On this basis, we introduce the StoneMLP model as the backbone, then describe a lightweight SparseFPN network and improve the detection head.
\subsection{Overall Framework}

In the ore image segmentation task, there is a problem of edge blur caused by the adhesiveness of ore images. 
Therefore, we propose a lightweight instance segmentation model with a point-based prediction strategy to address the edge problem, which utilizes a lightweight backbone and feature pyramid network (FPN) structure based on MLP to extract high-quality edge features. To further enhance the accuracy of edge prediction, we introduce an Edge Guidance Loss that dynamically aligns the predicted edge with the instance edge.

The OreNeXt is a two-stage instance segmentation model with a network structure as mentioned in~\textcolor{red}{Fig. \ref{Architecture}}. The model mainly consists of a lightweight MLP backbone, a SparseFPN, and a fine detection head. To better capture the low-level features, we designed a structure StoneMLP for extracting features of ore images as the backbone. By moving the feature information axially, information flows in different directions can be obtained, which helps to capture the local correlation between overlapping ores. The SparseFPN is used to extract hierarchical features. The feature maps are processed in the SparseMLP to efficiently obtain the global receptive field. Global information can assist in better describing semantic information. Furthermore, we use a lightweight segmentation head to generate coarse prediction features for each detected object. For the fine-grained feature from the CNN feature maps($S_i$), we perform bilinear interpolation by coordinate points on the feature maps to compute the feature vector. Then the fine-grained features that provide object detail information are combined with the coarse prediction features that provide global context information. The fused feature map is fed into the prediction head, making point-wise segmentation.

\subsection{StoneMLP}
We propose a backbone network StoneMLP based on the architecture of MLP. Compared to convolution and attention networks, the MLP structure has a lower inductive bias. It allows the model to learn solely from raw data, thereby making the network more concise. We propose a spatial shift method that enables the network to obtain different directions of information flow, achieving the same field of view as CNN and completing feature extraction. Instance segmentation of ore images necessitates the utilization of low-level features, including contours, textures, colors, and shapes. Our proposed spatial movement method effectively captures local correlations by prioritizing the extraction of these low-level features, thereby yielding more comprehensive edge information.

\begin{figure}[!t]
	\centering
	\includegraphics[width=3in]{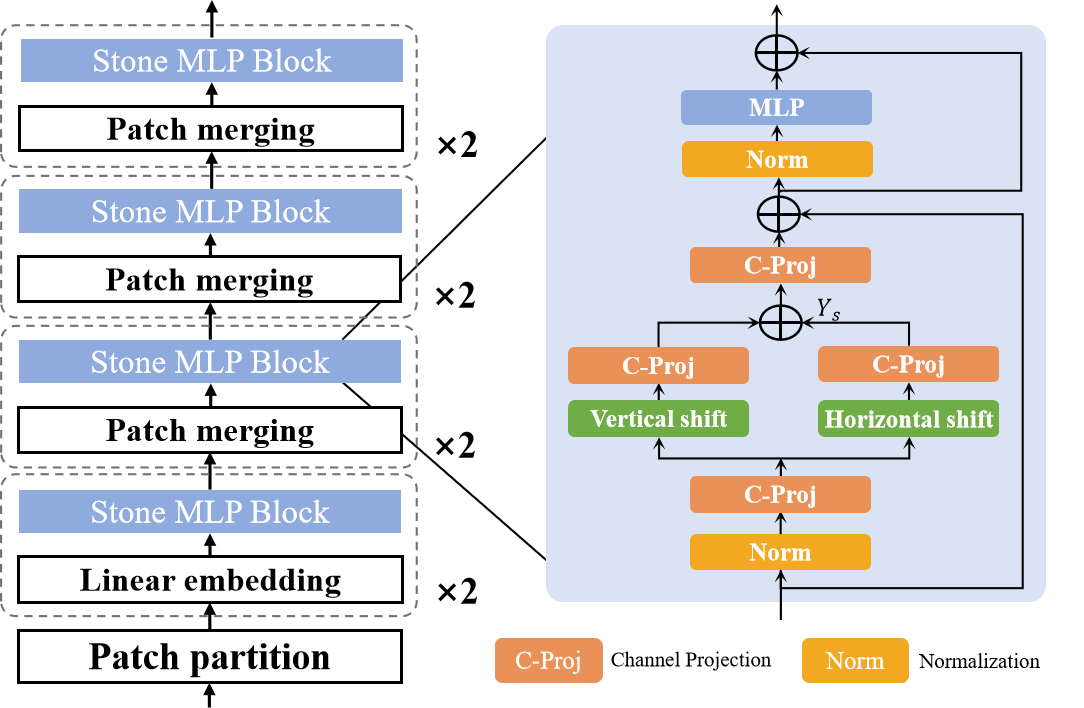}
	\caption{Architecture of StoneMLP. The proposed StoneMLP block mainly includes Norm, Pixel Shift operation, MLP, channel projection, and residual connection.}
	\label{StoneMLP}
\end{figure}

In \textcolor{red}{Fig.~\ref{StoneMLP}}, it shows the architecture of our StoneMLP model. For example, ResMLP~\cite{resmlp} has shown that self-attention is not the key factor of transformers for achieving excellent performance. With further research, many methods prove that self-attention modules can be replaced by MLP, Convolution, or other layers, indicating that the success of the transformer may come from the entire architecture design. Therefore,  we refer to Transformer~\cite{transformer} for the design of the overall structural framework to expect good results. It takes RGB images $I\in \mathbb{R}^{ 3\times H\times W}$ as input and then slices them into patches of $4 \times 4$ size. All tokens obtained at this stage are $48\times\frac{H}{4}\times\frac{W}{4}$.  
StoneMLP Block contains four fully connected layers, each corresponding to four channel projections for communicating specific location information. Channel projections, vertical movement, and horizontal movement are used to extract features in the Pixel Shift operation. The temporal and computational expenses associated with the axial movement are exceedingly minimal. The computational complexity is $\mathrm{\Omega(StoneMLP)}=4HWC^2$. 

\begin{figure}[!t]
	\centering  
	\subfigure[Horizontal shift]{
		\includegraphics[width=0.9\linewidth]{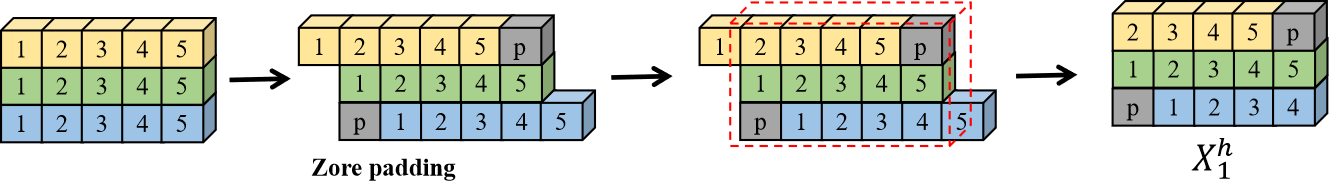}}
	\subfigure[Vertical shift]{
		\includegraphics[width=0.9\linewidth]{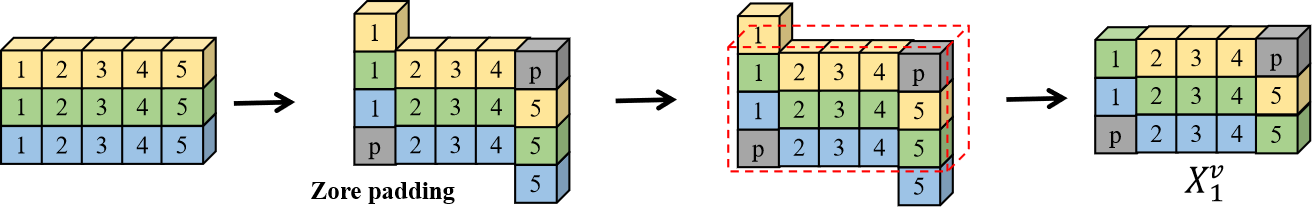}}
	%\quad
	\caption{The horizontal shift and vertical shift, where the arrows indicate the steps, and the number in each box is the index of the feature.}
  \label{shift}
\end{figure}

As illustrated in \textcolor{red}{Fig.~\ref{shift}}, we will explain the horizontal movement. The following channel projection will use the features that were extracted for the dashed box. The vertical movement operation is very similar to the horizontal movement. Combining the horizontal and vertical movements, the features in different positions are realigned on a channel. During the next channel projection operation, information from both directions is integrated to obtain the result after local communication. When the input $x$ and the displacement size are given, the output $Y_S$ is obtained.
\begin{equation}\label{output}
	Y_S=X^h \ W_c^h+X^v \ W_c^v,
\end{equation}
where $ W _ c ^ h$ and $W _ c ^ v$ denote the channel projection's learnable weights in the vertical and horizontal directions, $X ^ h = Concat ( X _ 1 ^ h,...., X _ s ^ h )$ and $X ^ v = Concat ( X _ 1 ^ v,... ., X _ s ^ v )$ represent the axial displacement.

\subsection{SparseFPN}

We propose a new SparseFPN for improving inference speed while maintaining accuracy. Traditional FPN structures generate multi-scale integrated features, due to the multi-level feature fusion, small targets are easily missed or misidentified. Therefore, we append two SparseMLP modules after traditional FPN and use a $1 \times 1$ convolution to output features. MLP can deal with more complex nonlinear relationships and learn more abstract feature representations.

\begin{figure}[!t]
	\centering  
	\subfigure[SparseMLP]{
		\includegraphics[width=0.9\linewidth]{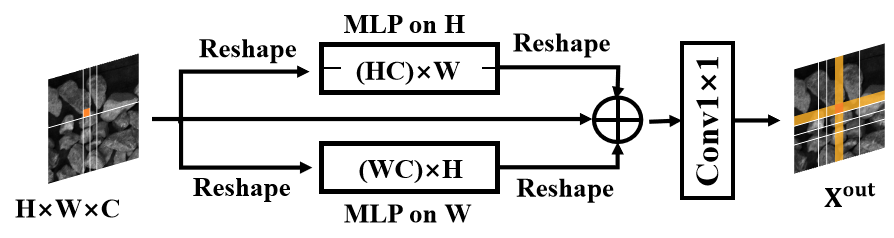}}\hspace{8mm}
       \subfigure[Receptive Field]
{
    \begin{minipage}[b]{.3\linewidth}
        \centering
        \includegraphics[width=0.95\linewidth]{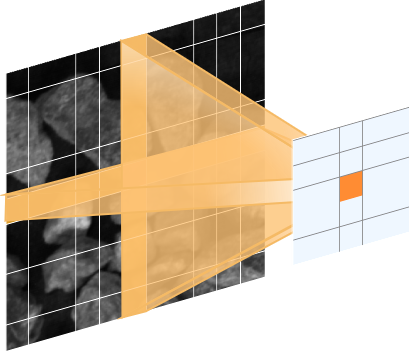}
    \end{minipage}\hspace{8mm}

 	\begin{minipage}[b]{.3\linewidth}
        \centering
        \includegraphics[width=0.95\linewidth]{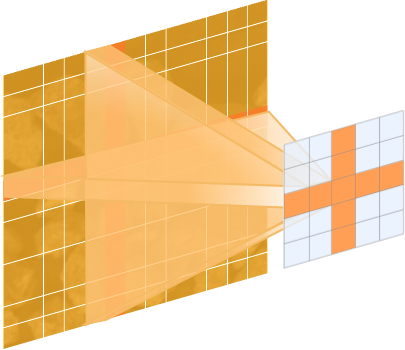}
    \end{minipage}
}

	%\quad
	\caption{The SparseMLP block consists of three branches: two are responsible for mixing information along the horizontal and vertical directions, respectively, and the other path is a constant mapping. (b) shows the receptive fields generated by two consecutive SparseMLPs.}

  \label{field}
\end{figure}

The structure of SparseMLP is demonstrated in~\textcolor{red}{Fig. \ref{field}(a)}, which is a module based on MLP with a parallel structure composed of three parts: \textbf{W} channel mapping, \textbf{H} channel mapping, and identity mapping. In the horizontal mixing path, features are reshaped and mixed information for each row. In the vertical mixing path, similar operations are applied. The three parallel branches are connected by channels, and feature fusion is achieved by weighted summation and $1 \times 1$ convolution, producing an output tensor $X^{out}=FC(concat(X_H\ ,\ X_W\ ,\ X))$ with the same dimensions as the input tensor. SparseMLP modules avoid the common overfitting problem affecting MLP-based models' performance. Each token only interacts directly with tokens in the same row or column, and each row and column can share the same projection weight. 
As shown in~\textcolor{red}{Fig. \ref{field}(b)}, a layer of SparseMLP will form a cross-shaped receptive field, and after two SparseMLPs, a global receptive field can be formed. That is to say, if this module is repeated twice, each token can accumulate information throughout the two-dimensional space, improving accuracy while reducing computational complexity $\mathrm{\Omega(SparseMLP)}=HWC(H+W)+3HWC^2$.

\subsection{Edge Guidance Loss}
To address the uncertainty resulting from the blurriness of edges in ore images, we use PointRend~\cite{pointrend} as the baseline and perform a recomputation of boundary hard pixels. We adopt the point selection and point-wise representation strategies to prioritize uncertain edge points for optimization of image segmentation for object edges. Additionally, we propose the Edge Guidance Loss function to improve the model performance. 
The loss can be divided into the following four parts:
\begin{equation}\label{Loss}
L_{EG}={\underbrace{L_{cls}^b\mathrm{+}\ \alpha L_{ploc}^b}_{L_{det}}}\mathrm{+}{\underbrace{\beta L_{coarse}^m\mathrm{+}\ L_{pmat}^m}_{L_{mask}}}.
\end{equation}
where $\alpha$ and $\beta $ represent the weight of $L_{ploc}^b$ and $L_{coarse}^m$, which is generally set to 0.5 and 1.0 respectively in the experiments. To improve the success rate of edge detection and segmentation, in addition to the classification loss and the coarse mask loss, we also introduce an Edge point Guidance match Loss and improve the target box offset loss to improve the positioning accuracy. 

In our method, $L _ { cls } ^ b$ and $L_{coarse}^m$ are defined as a standard cross entropy loss function with softmax activation. We treat the foreground points as positive samples and the other points as negative samples. The new regression localization loss $ L _ { ploc } ^ b$ is responsible for predicting the foreground score of the prediction point:
\begin{equation}\label{lossdet }
	L_{ploc}^b (R,R^\prime) = \frac{1}{n}\sum\limits_{k=1}^n\parallel(x_i,y_i)-(x_i^\prime,y_i^\prime)\parallel_2.
\end{equation} 
It uses the point-to-point distance prediction, input predicted point, and real border box. Moreover, it sorts the coordinates of test points in the x or y direction and obtains new coordinates. The distance between each point and four borders is calculated, and the boundary loss of each border point is obtained, which is finally obtained by weighted addition.

\begin{figure}[!t]
	\centering  
		\includegraphics[width=0.85\linewidth]{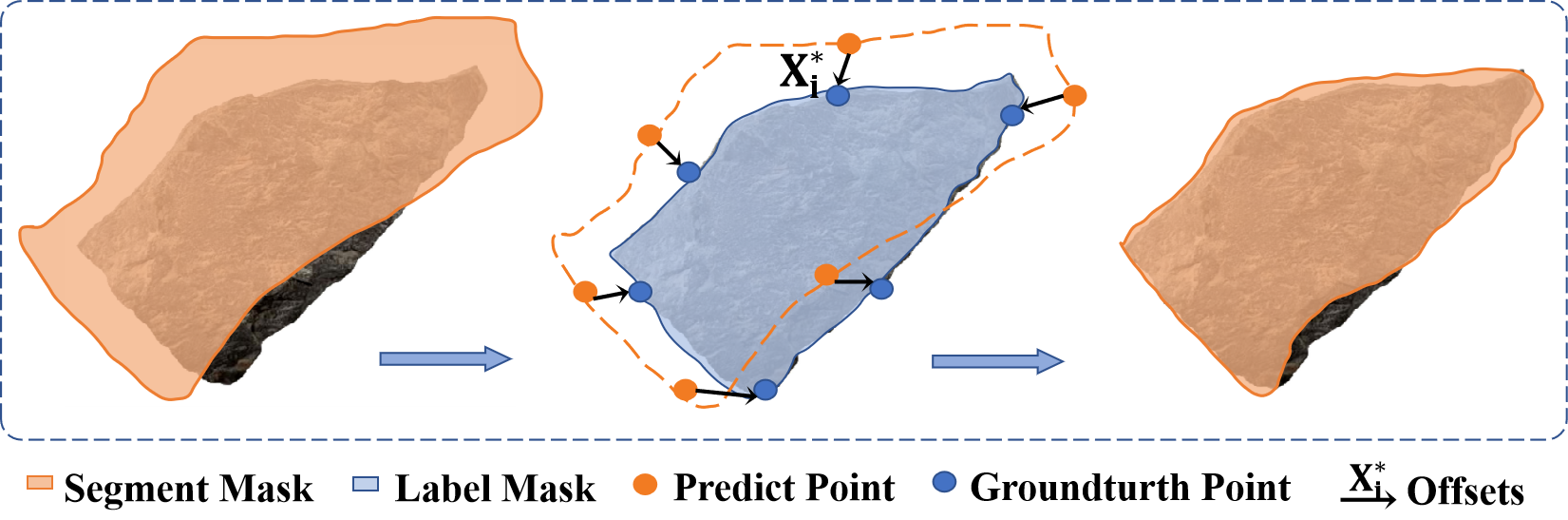}
	\\
 
	\caption{\textbf{Edge} point \textbf{Guidance} match \textbf{Loss}. The yellow points represent predicted vertices, while the blue points indicate labeled vertices. The arrows depict the path, representing the pairing relationship. Each prediction point is adjusted to the nearest point on the truth boundary.}
  \label{loss}
\end{figure}

The edge guidance match loss $ L _ { pmat } ^ m $ is the average of the traditional cross-entropy loss $L_{pcls}$ function and the edge point guidance match loss $L_{pg}$:
\begin{equation}\label{lossmask }
L_{pmat}^m=\frac{1}{2}(L_{pcls}+L_{pg}).
\end{equation}
In \textcolor{red}{Fig. \ref{loss}}, the $L_{pg}$ first calculates the distance between the prediction point $Pred_i^{in}$ and the real border point, finds the nearest target point $gt_{X}^{ipt}$ to the prediction point according to the distance, and interpolates the target points to a specified number of points using an interpolation function to obtain the offset difference $X_i^*$. It also increases the classification accuracy of the target and makes the target smoother.
\begin{equation}\label{x }
X_i^*=arg\min_{X} \parallel{Pred}_i^{in}-{gt}_{X}^{ipt}\parallel_2.
\end{equation}
Then, the predicted points ${ Pred}_i^{out}$ and the matched nearest interpolation label points ${ gt}_{X_i^*}^{ipt}$ are dynamically matched using the loss function smoothL1 to supervise the quality of boundary prediction. The edge point guidance match loss $L_{pg}$ can consider the smoothness and robustness of large error and small error, which is not affected by outliers.
\begin{equation}\label{point-match }
L_{pg}=\frac{1}{N}\sum_{k=1}^{n}\parallel{ Pred}_i^{out}-{ gt}_{X_i^*}^{ipt}\parallel_1.
\end{equation}

\section{Experiments}\label{sec:experiments}
In this section, we introduce the hardware environment for ore processing tasks and the datasets, evaluation metrics, and implementation details. Next, we conducted ablation research to evaluate the effectiveness of the design decisions. Eventually, we compare OreNeXt with the state-of-the-art methods for instance segmentation on the ore dataset.

\subsection{Hardware Environment}
\begin{figure}[!t]
\centering
\subfigure[Production site]
{
    \begin{minipage}[b]{.51\linewidth}
        \centering
        \includegraphics[scale=0.5]{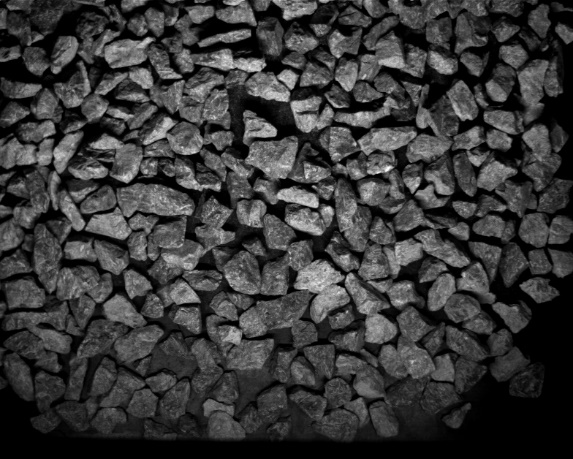} \\\vspace{0.4em} 
        \includegraphics[scale=0.355]{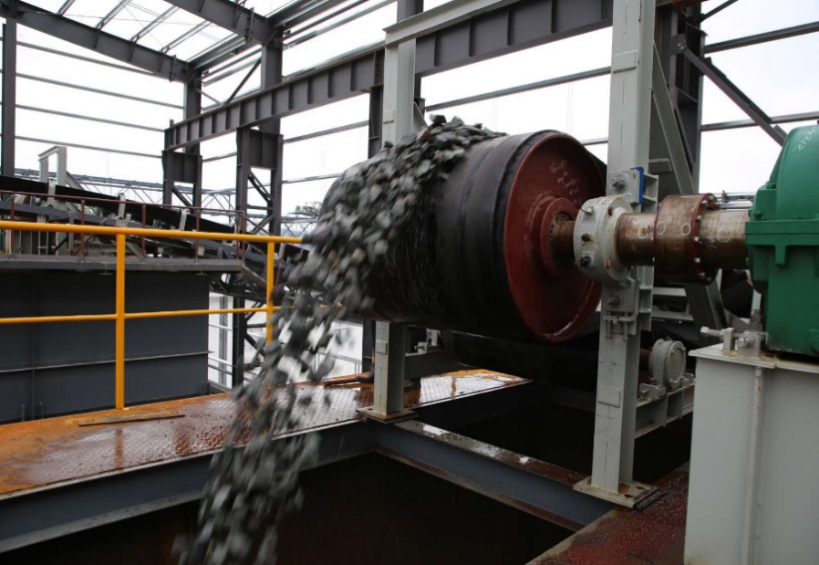}
    \end{minipage}
}\hspace{-2.5em}
\subfigure[Conveyor belt]
{
    \begin{minipage}[b]{.52\linewidth}
        \centering
        \includegraphics[scale=0.18]{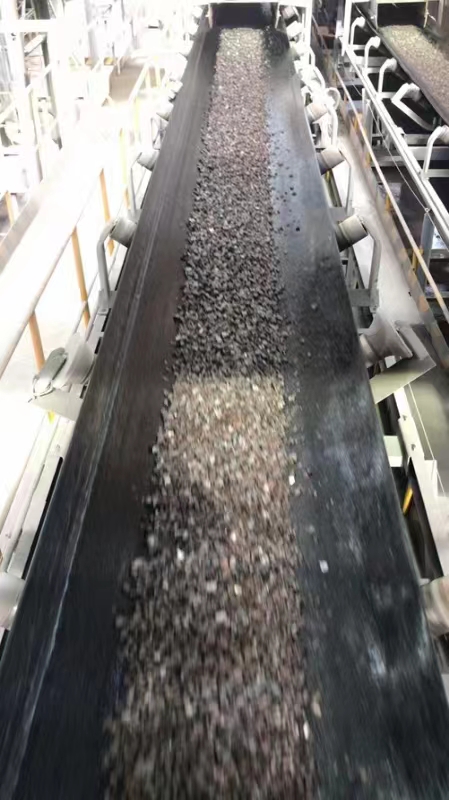}
    \end{minipage}
}
	\caption{Conveyor belt at the mine production site. The site environment is simple and the computing resources are poor. There is serious adhesion between the ores and the edges are blurred.}
\label{actualscenario}
\end{figure}

The actual scene of the ore processing tasks is depicted in~\textcolor{red}{Fig. \ref{actualscenario}}, revealing significant adhesion of the ore and blurred edges. The detection system is responsible for identifying the size indexes of ores on the conveyor belt, providing insights into the effectiveness of the upstream processes, and facilitating the identification of over-specification ores. Ore detection on the production line is performed through sampling inspection. The traditional manual screening involves the random selection of a set of ores as a sample to assess their compliance with required standards. The online detection method based on deep learning considers the complete ores present on the surface of the conveyor belt, as captured by an industrial camera, as a representative sample, so we only expect to segment independent and complete ore individuals to obtain their particle size information. Therefore, precise segmentation of the surface ores is crucial, particularly in accurately identifying the edges of overlapping ores.
It can be seen from Fig. \ref{actualscenario} that ore processing tasks are typically conducted outdoors, where hardware resources are limited, making the use of high-performance computing equipment impractical. Consequently, the model size becomes a critical factor in ore processing scenarios with constrained computing resources. We verify on the ORE dataset that compared with the current segmentation methods, our method achieves optimal accuracy while ensuring minimal memory usage.

\subsection{Experiments Setup}
\subsubsection{Datasets}
In the experiment, we apply ImageNet~\cite{imagenet} to pre-train the backbone model and train instance segmentation models using our created rock data set. For the ORE image dataset, we use 4,060 images for training and 1,060 images for validation. We supplement the data set in the following ways: first, we collect different scales of rock images using our experimental platform. Then, we change the positioning of rocks at different scales, such as sparse, thick, and so on. In addition, We break up large images into smaller ones that can be utilized to train networks. We segment images using a sliding window to further expand the datasets.%\footnote{https://drive.google.com/file/d/1eYkPHgDWULHind802P4tvy9l7lIQrpqk/view?pli=1.} 

\subsubsection{Evaluation Metrics}
To confirm the efficiency of the suggested method, seven evaluation indicators are used such as~$AP^{box}$,~$AP_{50}^{box}$,~$AP^{mask}$,~$AP_{50}^{mask}$, inference time (FPS), inference memory consumption, and model size. The first four indicators are used to evaluate the accuracy. AP stands for Average Precision, and the number after the AP represents the IoU(Intersection over Union) threshold, which refers to the degree of overlap between the predicted box and the ground truth box. 
In general, a higher value of $AP_{50}$ indicates a better performance of the algorithm. The latter three metrics evaluate whether a model can be deployed on low-cost hardware. The computational cost of the model is quantified by its inference time, and we aim to maximize this value. The utilization of memory during training and the model's size demonstrate its reliance on the hardware system, and we strive to minimize these factors while maintaining accuracy.
\subsubsection{Implementation Details}
OreNeXt is trained via back-propagation and the SGD (stochastic gradient descent) optimizer. For the ORE image dataset, all models shown in~\textcolor{red}{Table \ref{SOTA}} are trained for 12k iterations on an NVIDIA RTX2080Ti GPU. We set the initial learning rate to 0.001 cut it by 1/10 at 8k iterations and use a batch size of 8 for all ablation studies. The input images are changed during training to have short sides between 160 and 320 pixels and long sides of 320 pixels. The experiments on the MS COCO dataset, the pre-train of the StoneMLP, and SAM-related experiments are all trained on the NVIDIA RTX3090Ti GPU. To ensure fairness and reproducibility in the experiment, a fixed random seed is used, and the hyperparameters of all models are set according to the best values provided in their respective papers. The parameters that achieve the highest accuracy on the validation set are preserved for prediction.

\subsection{Comparison with State-of-the-Art Methods}
\subsubsection{Classic Segmentation}
To further demonstrate the supremacy of OreNeXt, we conduct a comparative analysis with state-of-the-art methods on the ORE dataset. Specifically, we compare the segmentation performance of a typical two-stage segmentation framework utilizing various CNN, Transformer, and MLP backbone networks. The results can be found in~\textcolor{red}{Table \ref{SOTA}}, which compares the segmentation accuracy, inference time, model size, and GPU memory usage.

\begin{table*}[!t]
\renewcommand{\arraystretch}{1}
  \centering
  \caption{ Comparison of accuracy with the state-of-the-art methods on ore image datasets. T-based: Transformer-based.}
    \begin{tabular}{c|lcccccccc}
    \toprule
\multicolumn{2}{c}{\multirow{1.5}[2]{*}{\textbf{Method}}} & \multirow{1.5}[2]{*}{\textbf{Backbone}} & \multirow{1.5}[2]{*}{$AP^{box}\uparrow$} & \multirow{1.5}[2]{*}{$AP^{box}_{50}\uparrow$} &  \multirow{1.5}[2]{*}{$AP^{mask}\uparrow$} & \multirow{1.5}[2]{*}{$AP^{mask}_{50}\uparrow$} &  \multirow{1.5}[2]{*}{\textbf{FPS$\uparrow$}} & \multirow{1.5}[2]{*}{\textbf{\makecell[c]{Model Size \\(MB)}$\downarrow$}} &\multirow{1.5}[2]{*}{\textbf{\makecell[c]{Inf.Memory\\(MB)}$\downarrow$}}  \\
\multicolumn{2}{c}{} &       &       &       &       &       &       &       &              \\
    \midrule
\multirow{12}[1]{*}{\begin{turn}{90}\emph{\textbf{CNN-based}}\end{turn}} & Mask RCNN\cite{maskrcnn}  & ResNet101 & 43.0    & 51.7    & 12.1  & 22.1   & 13  & 480   & 1876 \\
    & PointRend\cite{pointrend}  & ResNet101 & 37.4  & 54.0       & 21.0    & 43.8    &17       & 489   &3361  \\

    & YOLACT\cite{YOLACT}  & ResNet101 & 40.3  & 50.6    & 7.1   & 11.9  &24   & 410   & 7758 \\
    & Mask RCNN\cite{maskrcnn}  & ResNet50 & 42.8  & 51.7    & 12.4  & 22.6   & 17  & 334   & 1769 \\
    & PointRend\cite{pointrend}  & ResNet50 & 36.3  & 54.0     & 25.8  & 47.7  &   21    & 344   &2677  \\
    & Cascade Mask RCNN\cite{Cascade-RCNN} & ResNet50 & 41.6  & 51.7     & 36.4  & 48.3    &  16     & 587   &1947  \\
    
    & CondInst\cite{CondInst} & ResNet50 & 43.1  & 52.2    & 39.0  & 48.7   & 23  & 259 & 1713 \\
    % & CARAFE\cite{CARAFE}  & ResNet50 & 42.9  & 51.8   & 13.7  & 24.9  & 15    & 376   & 1873 \\
    & MS RCNN\cite{Mask-Scoring-RCNN}  & ResNet50 & 42.7  & 51.7    & 14.7  & 14.7    &16  & 428   & 1749 \\
    & Blend Mask\cite{BlendMask}  & ResNet50 & 44.1  &58.1 & 38.9  & 48.7    & 23  & 274   & 1641 \\
    & BoxInst\cite{BoxInst}  & ResNet50 & 44.3  & 56.9    & 34.8  & 48.7   & 23  & 261   & 1681 \\
    & SparseInst\cite{SparseInst}  & ResNet50 & -  & -    & 24.8 & 37.9  & 21  & 380   & 1711 \\

    % & Mask2Former\cite{mask2former}  & ResNet50 & 58.3  & 74.8    & 35.3 & 44.3  & 20  & 572   & 6098 \\
    &RTMDet\cite{lyu2022rtmdet}&CSPNeXt &38.7&	49.7&35.1&	43.4& 22&109&2013
\\
       \midrule
\multirow{6}[1]{*}{\begin{turn}{90}\emph{\textbf{T-based}}\end{turn}} & Mask RCNN\cite{maskrcnn}  & Swin-T & 38.3  & 52.0   & 27.7  & 46.5 & 19 & 542   & 1995 \\
    & PointRend\cite{pointrend}  & Swin-T & 30.6  & 55.4 & 32.8   & 44.7    &    16   & 371   &  2631\\
    & Cascade Mask RCNN\cite{Cascade-RCNN} & Swin-T & 42.0  & 50.7  & 38.5  & 48.5  &14  & 920   &2063  \\
    & Mask2Former\cite{mask2former} & Swin-T & 55.7   & 71.3  & 35.3  & 44.3  &20  & 572   &6098  \\
    & Mask RCNN\cite{maskrcnn}  & PVTv2   & 39.6  & 51.4      & 27.3  & 44.8    & 24& 263   & 1943 \\
    & PointRend\cite{pointrend}  & PVTv2   & 39.8  & 51.8   & 16.1  & 43.0    &  19     & 185   & 2501 \\
    & Cascade Mask RCNN\cite{Cascade-RCNN} & PVTv2   & 42.7  & 50.9    &39.0   & 48.7  &      17 & 641   &1907  \\
    \midrule
    \multirow{6}[1]{*}{\begin{turn}{90}\emph{\textbf{MLP-based}}\end{turn}} & Mask RCNN\cite{maskrcnn}  & ASMLP & 35.9  & 49.2    & 21.1  & 38.5   &21 & 542 &1811\\
    & PointRend\cite{pointrend}  & ASMLP & 36.6  & 52.6   & 27.0    & 41.1  &     20  & 371&2389 \\
    & Cascade Mask RCNN\cite{Cascade-RCNN} & ASMLP &42.4 & 50.7    & 38.7 & 48.5  & 19& 613   & 2043 \\
    & Mask RCNN\cite{maskrcnn} & CycleMLP-B1 & 36.9  & 51.7  & 29.1  & 45.2  &  26 & 104   &  1673\\
    & PointRend\cite{pointrend}  & CycleMLP-B1 & 25.0    & 50.2   & 20.0    & 39.8  &26 & 109   &1843  \\
    & \textbf{OreNeXt(Ours)} & \textbf{StoneMLP} & 38.2 & 60.4 & 33.8 & 48.9 &  27  & 73 &1931  \\
    \bottomrule 
    \end{tabular}
  \label{SOTA}%
\end{table*}%

Firstly, the networks based on CNN have higher accuracy but are slower and have larger model sizes than ours. Since the transformer structure has the largest model capacity, the transformer-based networks have a larger model size, while the MLP-based networks have a smaller model size but worse accuracy. Secondly, Our proposed OreNext network exceeds the best CNN and Transformer algorithms both by 0.2 on~$AP_{50}^{mask}$ metrics, while also achieving high accuracy on~$AP_{50}^{box}$, due to our novel approach of using a Sparse FPN that balances local and global information with two MLP structures, as well as an Edge Guidance Loss based on a point processing strategy to improve accuracy. Recent methods like Mask2Former~\cite{mask2former} based on SwinTransformer~\cite{swint} have achieved impressive results on the~$AP^{box}$ metric, surpassing the average performance, while its corresponding model size is $6\times$ larger than our OreNeXt. A small model size is crucial in ore processing tasks with limited computational resources. Therefore, models that prioritize box accuracy at the expense of model size are not suitable for field ore tasks. Furthermore, OreNeXt also outperforms other lightweight frameworks significantly in inference time, model size, and GPU memory usage, thanks to our lightweight backbone StoneMLP, which is even smaller than other lightweight MLP-based backbone networks such as CycleMLP. OreNeXt strikes a balance between accuracy and computational cost and is more suitable for ore segmentation tasks. Finally, the performance of Swin-T and ASMLP used as the backbone network of PointRend is similar to that of OreNext in terms of accuracy and model size, which supports the idea that the token mixer module of the transformer model is not the key to its excellent performance but the entire transformer architecture. Therefore, replacing the most time-consuming attention of the transformer with the StoneMLP network can also achieve excellent performance. 

\subsubsection{Segment Anything Model}
% Table generated by Excel2LaTeX from sheet 'pyt'
\begin{table}[!t]
  \centering
  \renewcommand{\arraystretch}{1}
  \caption{The comparison of OreNeXt with SAM-based instance segmentation methods, $\star$ means only validation}
\resizebox{\columnwidth}{!}{
    \begin{tabular}{lcccc}
    \toprule
    \textbf{Method} & $ {AP^{box}_{50}}$ & $ {AP^{mask}_{50}}$ & \textbf{FPS} & \textbf{\makecell[c]{Model Size\\ (MB)}} \\
    \midrule
    SAM\cite{segmentanything} & 42.2  & 44.7  &  3     & 609 \\
    RSPrompter\cite{rsprompter} & 42.7  & 41.6  & 2      & 631 \\
    $FastSAM^{\star}$\cite{fastsam} & 38.0    &  37.2     &  7     & 150 \\
    OreNeXt(Ours)  & 60.4  & 48.9  & 27    & 73 \\
    \bottomrule
    \end{tabular}
  \label{sam}}
\end{table}%
By finding foundational models that exhibit high performance in the fields of NLP and CV, the foundation Segment Anything Model (SAM)~\cite{segmentanything} proposed by Meta AI exhibits remarkable generalization capabilities. SAM is becoming a foundation step for many visual tasks, like image editing and remote sensing image segmentation. In the ore processing task, we compare the existing instance segmentation methods based on the SAM foundation model with our method, as shown in~\textcolor{red}{Table \ref{sam}}. As a category-agnostic instance segmentation method, SAM does not show its superiority in single-class ore image segmentation. Its segmentation accuracy is average, but the model size exceeds most traditional segmentation methods. In single-task industrial application scenarios, SAM not only fails to play its advantages of strong generalization ability, but its huge computation costs prevent it from wider applications in industry scenarios. At present, the traditional instance segmentation method is still the best choice after balancing accuracy and computational complexity in industrial applications. In the future, the goal is to achieve high-performance compression of large models and develop comprehensive industrial foundation modes capable of addressing multiple scenarios and tasks.

\begin{figure*}[!t]
	\centering
	
	\subfigure[Input]{
		\begin{minipage}[c]{0.13\linewidth}
			\includegraphics[width=0.93in]{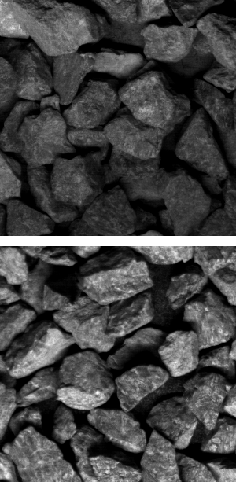}\vspace{0.4em} 
		\end{minipage}
	}\hspace{0.1em}
    \subfigure[GroundTruth]{
		\begin{minipage}[c]{0.13\linewidth}
			\includegraphics[width=0.93in]{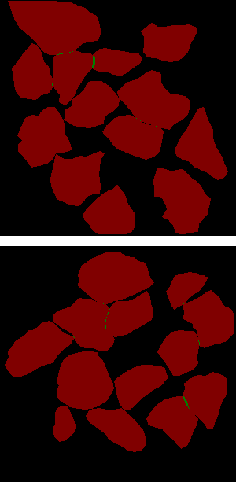}\vspace{0.4em} 
		\end{minipage}
    }\hspace{0.1em}
     	\subfigure[SagmentAnything]{
		\begin{minipage}[c]{0.13\linewidth}
			\includegraphics[width=0.93in]{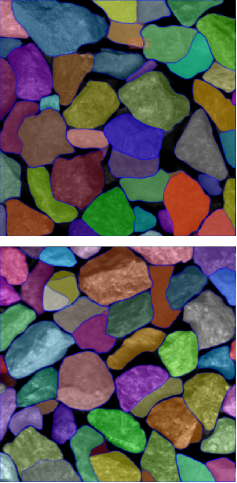}\vspace{0.4em} 
		\end{minipage}
	}\hspace{0.1em}
	\subfigure[PointRend]{
		\begin{minipage}[c]{0.13\linewidth}
			\includegraphics[width=0.93in]{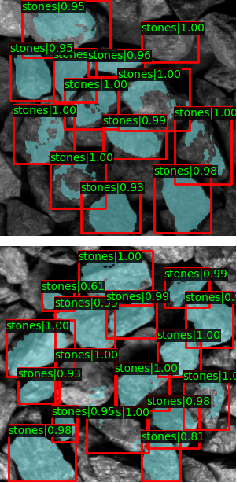}\vspace{0.4em} 
		\end{minipage}
	}\hspace{0.1em}
    \subfigure[YOLACT]{
		\begin{minipage}[c]{0.13\linewidth}
			\includegraphics[width=0.93in]{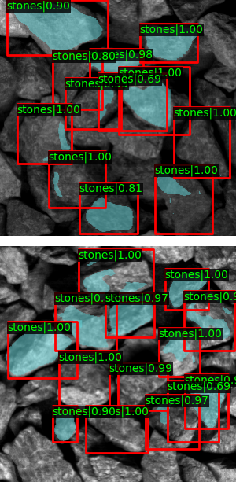}\vspace{0.4em} 
		\end{minipage}
	}\hspace{0.1em}
	\subfigure[OreNeXt]{
		\begin{minipage}[c]{0.13\linewidth}
			\includegraphics[width=0.93in]{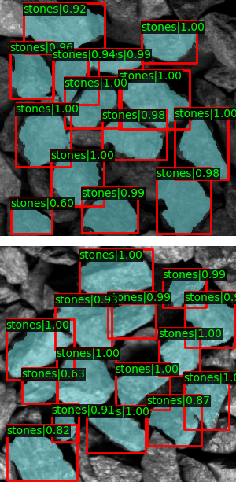}\vspace{0.4em} 	
		\end{minipage}
	}
	\caption{The visual results of ore images processed by different methods. It can be seen that our method (f) has the best mask quality, and the foundation model (c) gets too much information about incomplete ores, which is unsuitable for actual production.}

  \label{sota}
\end{figure*}

The visualization of various results from the instance segmentation method in ore images is depicted in \textcolor{red}{Fig.~\ref{sota}}. It can be observed that our method is significantly better than other methods in mask quality as well as edge sharpness, and the missed detection rate is also at a low level. Compared to the foundation large model, our method can detect as many complete ore images as possible. It can obtain the granularity information required for industrial production and meet the needs of sampling inspection.

\subsection{Ablation Study}
\subsubsection{StoneMLP}
To demonstrate the effectiveness of our proposed backbone network, we verified the accuracy of this backbone network by comparing various backbone changes. In~\textcolor{red}{Table \ref{backbone}}, StoneMLP outperforms the other backbone networks in three indicators of $AP^{box}$, $AP^{mask}$, $AP_{50}^{mask}$. The index $AP_{50}^{box}$ of evaluating detection bounding box is 60.4, which is slightly lower than the highest value of ResNet101 in 61.9, while the model size of 73M is 6.5$\times$ smaller than ResNet101. The small model size is also vital in ore processing tasks with limited computational resources. Since StoneMLP is designed for feature extraction of ore low-level edge information, it greatly outperforms the lightweight backbone CycleMLP with 99M, which is also based on MLP in terms of accuracy across the board. In order to verify the general effectiveness of StoneMLP, we conduct experiments on the MS COCO~\cite{coco} dataset, and the results in~\textcolor{red}{Table \ref{coco}} show that the accuracy is significantly improved compared with the baseline.

\begin{table}[!t]
\renewcommand{\arraystretch}{1}
  \centering
  \caption{The segmentation results of different backbones in OreNeXt}
   \resizebox{\columnwidth}{!}{
   \setlength{\tabcolsep}{2pt}{
    \begin{tabular}{l|ccccc}
    \toprule
    \textbf{\makecell[l]{Backbone}} & \textbf{$ {AP^{box}}$} & ${AP^{box}_{50}}$ & \textbf{$ {AP^{mask}}$}  & \textbf{$ {AP^{mask}_{50}}$}  & \textbf{\makecell[c]{Model Size\\ (MB)}} \\
    \midrule

    ResNet50\cite{ResNet} & 36.8    &60.8  & 31.1   & 47.4   & 329 \\
    ResNet101\cite{ResNet} & 34.0        & \textbf{61.9}    & 29.9    &48.3       & 474 \\
    RegNet\cite{RegNet} & 36.1  & 57.1    & 29.2   & 47.4      & 255 \\
    HRNet\cite{hrnet}& 36.7  & 61.8    & 25.7   & 25.1      & 229 \\
    PVTv2\cite{pvt}  &  36.1 &  61.3  & 22.5  & 47.7      &178  \\
    Swin-T\cite{swint} & 30.3   & 61.2 & 11.8    & 25.8   & 220 \\
    
    ASMLP\cite{asmlp} & 36.4   & 58.9  & 30.5    & 46.8   & 214 \\
    CycleMLP-B1\cite{cyclemlp} & 26.6    &59.3 & 22.4   & 43.8 & 99 \\
    StoneMLP(ours) & \textbf{38.2}   & 60.4 & \textbf{33.8}  & \textbf{48.9}  &\textbf{73}\\
    \bottomrule
    \end{tabular}}}
  \label{backbone}%
\end{table}%

\begin{table}[!t]
\renewcommand{\arraystretch}{1}
  \centering
  \caption{Comparison crucial components on MS COCO datasets}
\resizebox{\columnwidth}{!}{
    \begin{tabular}{l|cccc}
    \toprule
          & $ {AP^{box}}$ & $ {AP^{box}_{50}}$ & $ {AP^{mask}}$& $ {AP^{mask}_{50}}$ \\
    \midrule
    Baseline & 21.6  & 37.4  & 20.3  & 34.5 \\
    EGLoss & $19.9_{~\textcolor{blue}{\tiny-1.7}}$ & $39.3_{~\textcolor{red}{\tiny+1.9}}$ & $21.1_{~\textcolor{red}{\tiny+0.8}}$ & $36.8_{~\textcolor{red}{\tiny+2.3}}$ \\
    StoneMLP & $25.1_{~\textcolor{red}{\tiny+3.5}}$  & $41.2_{~\textcolor{red}{\tiny+3.8}}$  & $23.6_{~\textcolor{red}{\tiny+3.3}}$  & $38.8_{~\textcolor{red}{\tiny+4.3}}$\\

    \bottomrule
    \end{tabular}}
  \label{coco}%
\end{table}%

\begin{table}[!t]

\renewcommand{\arraystretch}{1}
  \centering
  \caption{Comparison of using different MLP in SparseFPN}
\resizebox{\columnwidth}{!}{
    \begin{tabular}{lcccc}
    \toprule
    \textbf{Method} & $ {AP^{box}_{50}}$ & $ {AP^{mask}_{50}}$ & \textbf{FPS} & \textbf{\makecell[c]{Model Size\\ (MB)}} \\
    \midrule
    MLP\cite{swint}   & 51.6  & 39.5  & 25 & 84 \\
    CycleMLP\cite{cyclemlp} & 50.5  & 37.7  & 23 & 83 \\
    SparseMLP(ours)  & \textbf{53.0} & \textbf{48.5} & \textbf{27} & \textbf{77} \\
    \bottomrule
    \end{tabular}}
  \label{fc}%
\end{table}%
\subsubsection{SparseFPN}
To improve detection accuracy while maintaining light weight, we designed an MLP module SparseMLP to enhance the receptive field by mixing information and fusing features through channel mapping. The module can be implemented using a simple MLP structure. In~\textcolor{red}{Table \ref{fc}}, we compare the effectiveness of different MLP structures. Experimental results show that connecting two SparseMLP after FPN can improve the receptive field of images, reduce the parameter count to avoid overfitting, and promote multi-stage processing in the pyramid architecture. The SparseFPN can strengthen the small object information that is easy to misjudge while maintaining sufficient semantic information.

\subsubsection{Loss Function}

To verify the robustness of EGLoss to the network, we compared the baseline loss with the newly proposed edge guidance loss and found that the ~$AP_{50}^{box}$ and~$AP_{50}^{mask}$ values were significantly improved while the model size remained constant. Additionally, we compared the accuracy stability before and after replacing the loss and observed that the~$AP_{50}^{mask}$ value became relatively stable after replacing the loss. In~\textcolor{red}{Table~\ref{coco}}, experiments on the public dataset MS COCO\cite{coco} show that EGLoss achieves good results on $AP_{50}^{mask}$, which is an indicator for evaluating mask quality.

\begin{table}[!t]
\renewcommand{\arraystretch}{1}
  \centering
  \caption{Effects of the components of the proposed method. SM: StoneMLP, SF: SparseFPN, EL:EGLoss}
    \begin{tabular}{rrrrcccc}
    \toprule
    \multicolumn{1}{c}{\textbf{Baseline}} & \multicolumn{1}{l}{\textbf{SM}} & \multicolumn{1}{l}{\textbf{SF}} & \multicolumn{1}{l}{\textbf{EL}} & $ {AP^{box}_{50}}$ & $ {AP^{mask}_{50}}$ & \textbf{\makecell[c]{Model Size\\ (MB)}}\\
    \midrule
       \multicolumn{1}{c}{\ding{51} }  &       &       &       & 54    & 47.7     & 344 \\
         \multicolumn{1}{c}{\ding{51} } & \ding{51}       &       &       & 51.7  & 47.8    & 85 \\
         \multicolumn{1}{c}{\ding{51} } &  \ding{51}      & \ding{51}       &       & 53.9  & 48.4     & 77 \\
         \multicolumn{1}{c}{\ding{51} } &  \ding{51}      & \ding{51}       &    \ding{51}    & \textbf{60.4} & \textbf{48.9}   & \textbf{73} \\
    \bottomrule
    \end{tabular}
  \label{abliation}%
\end{table}%

\subsubsection{Module Breakdown Analysis}

We conducted ablation studies, as presented in~\textcolor{red}{Table \ref{abliation}}, to investigate the specific contributions of each module in OreNeXt. StoneMLP incorporates shifting operations to capture local information, resulting in a slight performance improvement while significantly reducing the number of parameters. The model size decreases from 344M to 95M. Next, the SparseMLP blocks are introduced into FPN, helping extract small features and improve accuracy. Finally, the Loss function is replaced with the EGLoss, leading to a significant improvement in accuracy from 53.9 to 60.4 in the $AP^{box}_{50}$ and from 48.4 to 48.9 in the $AP^{mask}_{50}$.

\subsubsection{Hyperparameter}
To explore the influence of hyperparameter settings on network performance, we conducted ablation experiments on the weight of the loss function and the number of channels in the neck. We have reduced the model size in SpraseFPN by changing the convolution size and reducing the number of channels. The results are depicted in~\textcolor{red}{Table~\ref{channel}}, and the model size gradually lowers as the number of channels reduces from 256 to 32. The highest value of accuracy is achieved when the number of channels is 64, which are 54.7 in~$AP_{50}^{box}$ and 47.6 in~$AP_{50}^{mask}$ respectively. The channels of the feature map contain the feature information, and the number of channels should be reasonably designed according to the feature complexity of the detected object. Reducing the number of channels can improve accuracy due to the ORE dataset having fewer shots and features than the public dataset. When the number of channels is reduced from 64 to 32, the accuracy does not increase, which proves that the feature information will also be lost when the number of channels is too small. A channel number of 64 increases the model size almost negligible compared to a channel number of 32.
These results indicate that a channel of 64 and $1 \times 1$ convolution is the optimal choice for ore segmentation.

Parameter $\beta $ in (\ref{Loss}) reflects the significance of $L_{coarse}^m$ in training. We perform ablation experiments on the MS COCO dataset and ORE dataset, respectively, and the results are reflected in~\textcolor{red}{Table~\ref{Weight}}. It can be seen that a smaller $\beta $ value can achieve the highest detection accuracy due to the single category of the ore dataset, while for the MS COCO dataset with rich target categories, a larger weight is needed to make the model not ignore the sample features of some categories.

\begin{table}[!t]
\small
  \renewcommand{\arraystretch}{1}
  \centering
  \caption{Comparisons of different channels in SparseFPN}

\resizebox{\columnwidth}{!}{
   \setlength{\tabcolsep}{6pt}{
    \begin{tabular}{cccccc}
    \toprule
    \textbf{Channels} & \textbf{Convs.} & ${AP^{box}_{50}}$ & $ {AP^{mask}_{50}}$ & \textbf{FPS} & \textbf{\makecell[c]{Model Size\\ (MB)}} \\
    \toprule
    \multirow{1.5}[2]{*}{32} & 1x1   & 52.2  & 46.1  &25  & 77 \\
          & 3x3   &     53.6	
  &   42.6    &   24    &  79\\
    \midrule
    \multirow{1.5}[2]{*}{64} & \textbf{1x1}   & \textbf{54.7}& \textbf{47.6}  & \textbf{26} & 80 \\
          & 3x3   & 52.6  & 46.7  & 24 & 88 \\
    \midrule
    \multirow{1.5}[2]{*}{96} & 1x1   & 53.2  & 45.6  & 25  & 90 \\
          & 3x3   & 52.9  & 44.3  & 24 & 98 \\
    \midrule
    \multirow{1.5}[2]{*}{128} & 1x1   & 52.2  & 46.4  & 26 & 98 \\
          & 3x3   & 54.6  & 46.5  & 23 & 109 \\
    \midrule
    \multirow{1.5}[2]{*}{256} & 1x1   & 52.0    & 44.9  & 23 & 130 \\
          & 3x3   & 52.7  & 44.3  & 22 & 161 \\
    \bottomrule
    \end{tabular}}}
  \label{channel}
\end{table}%

\begin{table}[!t]
\small
\renewcommand{\arraystretch}{1}

  \centering
  \caption{Validation of $L_{coarse}^m$ in the different datasets. NC means cannot converge}
    \begin{tabular}{c|ccc|ccc}
        \toprule
          & \multicolumn{3}{c|}{ORE} & \multicolumn{3}{c}{MS COCO} \\
    \midrule
   $\beta $    & 0.5   & 1     & 1.5   & 1.5     & 2.5     & 3\\
    $ {AP^{mask}_{50}}$   & 45.5  & 48.9  & 46.4  & 30.8  & 36.8  & NC \\
    \bottomrule
    \end{tabular}
  \label{Weight}%
\end{table}%

\section{Conclusion}\label{sec:conclusion}
This study presents OreNeXt, a lightweight and efficient instance segmentation network designed for ore image segmentation. The network specifically addresses the challenging issue of edge blurring that results from ore overlap.
OreNeXt is an MLP-based architecture that leverages two essential components to enhance low-level edge features. Firstly, it incorporates a StoneMLP backbone network structure that facilitates spatial information interaction through shift operations. Secondly, it incorporates a SparseFPN that effectively balances global and local information to improve accuracy. Additionally, a point-guided loss function is employed to enhance the clarity of edge segmentation.
When assessed using the ore dataset, our model demonstrated superior performance in ore segmentation compared to other existing methods while also exhibiting faster inference time, decreased computational complexity, and a reduction in the number of parameters required.
For future work, we will focus on further improving edge accuracy and increasing the model's speed. We will also explore edge segmentation algorithms that can be extended to other industrial areas.

\bibliographystyle{ieeetr}
\bibliography{IEEEabrv,BIB_xx-TIE-xxxx} %IEEEabrv instead of IEEEfull

\end{document}